\pdfoutput=1
\documentclass[11pt]{article}

\usepackage[]{ACL2023}
\usepackage{times}
\usepackage{latexsym}
\usepackage{tabularx}
\usepackage{multirow}
\usepackage{todonotes}
\usepackage{booktabs}
\usepackage{xcolor,colortbl}
\definecolor{Gray}{gray}{0.85}
\newcolumntype{g}{>{\columncolor{Gray}}c}
\usepackage{graphicx}
\usepackage{amssymb} 
\usepackage{xspace}

\usepackage[T1]{fontenc}
\usepackage[utf8]{inputenc}

\usepackage{microtype}

\usepackage{inconsolata}
\usepackage{todonotes}

\newcommand{\tokO}{\texttt{TOK-O}\xspace}
\newcommand{\tokS}{\texttt{TOK-S}\xspace}
\newcommand{\tokX}{\texttt{TOK-X}\xspace}

\newcommand{\tokSpace}{\textsc{Space}\xspace}
\newcommand{\tokDynamic}{\textsc{Dynamic}\xspace}
\newcommand{\tokTruncate}{\textsc{Truncation}\xspace}

\newcommand{\mlmFull}{{MLM-FULL}\xspace}
\newcommand{\mlmAdapt}{{MLM-ADAPT}\xspace}
\newcommand{\mlmEmb}{{MLM-EMB}\xspace}
\newcommand{\mlmEmbTok}{{MLM-EMBTOK}\xspace}
\newcommand{\mlmEmbTokS}{{MLM-EMBTOK-S}\xspace}
\newcommand{\mlmEmbTokX}{{MLM-EMBTOK-X}\xspace}
\newcommand{\mlmEmbs}{{MLM-EMBs}\xspace}

\newcommand{\mlmEmbsTokX}{{MLM-EMBTOKs-X}\xspace}

\newcommand{\TADA}{TADA\xspace}

\makeatletter
\newcommand{\srcsize}{\@setfontsize{\srcsize}{5pt}{5pt}}
\makeatother

\makeatletter
\newcommand{\mathsize}{\@setfontsize{\mathsize}{0.5pt}{0.5pt}}
\makeatother

\title{META-T: Robust Multi-Domain Meta Embeddings for Transformers}
\title{Transformer-Based Multi-Domain Meta-Embeddings for Robust Representation Learning}
\title{Next-Level Domain Adaptation: \\Enhancing Multi-Domain Meta Embeddings with Transformers}
\title{Next-Level Domain Adaptation: \\Robust Domain-Aware Input Representations with Transformers} % another version from Chia-Chien
\title{Next-Level Domain Adaptation: \\Domain-Aware and Multi-Domain Input Representations for Transformers}
\title{TADA: Efficient Task-Agnostic Domain Adaptation for Transformers}
\author{Chia-Chien Hung\textsuperscript{1, 2, 3}\thanks{$^\ast$Research work conducted during internship at Bosch Center for Artificial Intelligence.}\hspace{0.3em}, Lukas Lange\textsuperscript{3}, Jannik Str\"otgen\textsuperscript{3, 4}\\
\textsuperscript{1}NEC Laboratories Europe GmbH, Heidelberg, Germany \\
  \textsuperscript{2}Data and Web Science Group, University of Mannheim, Germany \\
  \textsuperscript{3}Bosch Center for Artificial Intelligence, Renningen, Germany \\
  \textsuperscript{4}Karlsruhe University of Applied Sciences, Karlsruhe, Germany \\
  \texttt{Chia-Chien.Hung@neclab.eu} \\
   \texttt{Lukas.Lange@de.bosch.com} \hspace{2em}
   \texttt{jannik.stroetgen@h-ka.de}}

\begin{document}
\maketitle

\begin{abstract}
Intermediate training of pre-trained trans\-former-based language models 
%(PTLMs) 
on domain-specific data leads to substantial gains for downstream tasks.
To increase efficiency and prevent catastrophic forgetting alleviated from full domain-adaptive pre-training,
%(i.e., updating all PTLM parameters), 
approaches such as adapters have been developed. 
However, these require additional parameters for each layer, and are criticized for their limited expressiveness. 
In this work, we introduce \TADA, a novel task-agnostic domain adaptation method which is modular, parameter-efficient, and thus, data-efficient. 
Within \TADA, we retrain the embeddings to learn domain-aware input representations and tokenizers for the transformer encoder, while freezing all other parameters of the model. Then, task-specific fine-tuning is performed.
We further conduct experiments with meta-embeddings and newly introduced meta-tokenizers, 
%utilizing multiple domain-specialized embeddings and tokenizers, respectively, 
resulting in one model per task in multi-domain use cases.
Our broad evaluation in 4 downstream tasks for 14 domains across single- and multi-domain setups and high- and low-resource scenarios reveals that \TADA is an effective and efficient alternative to full domain-adaptive pre-training and adapters for domain adaptation, while not introducing additional parameters or complex training steps.

% \textcolor{cyan}{TODO add meta-tokenizers in abstract and intro}
% \textcolor{cyan}{TODO2: Check all table/figure captions}
\end{abstract}

\section{Introduction}
%start from transformers
%encode domain knowledge into transformers
%two-step finetuning: MLM + training on downstream tasks
%full MLM specialization is not necessary: adapter, sparse finetuning
%proposal: light-weight method: update only the embeddings of transformers (portable, could be used as the domain-specific embeddings with contextual representation, same size)
%merge domain-speciaized embeddings in multi-domain scenario
\begin{figure*}[t]
	\centering
    \includegraphics[trim={0.3cm 0.7cm 1.2cm 4.7cm},clip,width=0.98\textwidth]{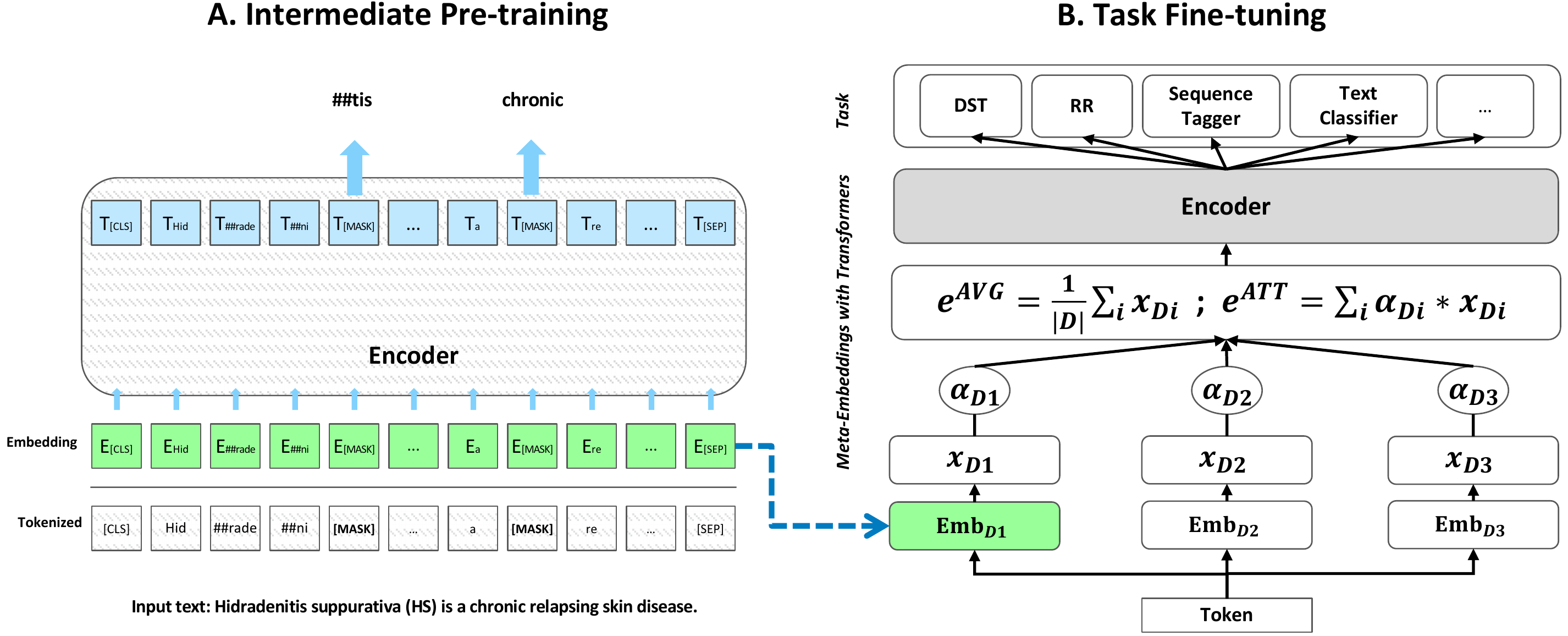}
    \vspace{-0.6em}
    \caption{Overview of the \TADA framework consisting of two steps. 
    Part A: Domain specialization is performed via embedding-based domain-adaptive intermediate pre-training %, operating on a small portion of in-domain text corpora 
    with Masked Language Modeling (MLM) objective on in-domain data. 
    Part B: The domain-specialized embeddings are then fine-tuned for downstream tasks in single- or multi-domain scenarios with two meta-embeddings methods: average (AVG) and attention-based (ATT).}
    \vspace{-1.2em}
    \label{fig:TADA}
\end{figure*}
%\vspace{-0.2em}
Pre-trained language models~\citep{radford2018improving, devlin-etal-2019-bert} utilizing transformers~\citep{vaswani-etal-2017-transformer} have emerged as a key technology for achieving impressive gains in a wide variety of natural language processing (NLP) tasks. However, these pre-trained transformer-based language models (PTLMs) are trained on massive and heterogeneous corpora with a focus on generalizability without addressing particular domain-specific concerns. 
In practice, the absence of such domain-relevant information can severely hurt performance in downstream applications as shown in numerous studies \citep[i.a.,][]{zhu2009introduction, ruder-plank-2018-strong,friedrich-etal-2020-sofc}. 

To impart useful domain knowledge, two main methods of domain adaptation leveraging transformers have emerged: 
(1)~\textit{Massive pre-training from scratch}~\citep{beltagy-etal-2019-scibert, wu-etal-2020-tod} relies on large-scale domain-specific corpora incorporating various self-supervised objectives during pre-training. However, the extensive training process is time- and resource-inefficient, as it requires a large collection of (un)labeled domain-specialized corpora and massive computational power. 
(2)~\textit{Domain-adaptive intermediate pre-training}~\citep{gururangan-etal-2020-dont} is considered more light-weight, as it requires only a small amount of in-domain data and fewer epochs continually training on the PTLM from a previous checkpoint. However, \textit{fully pre-training} the model (i.e., updating all PTLM parameters) may result in catastrophic forgetting and interference~\citep{mccloskey1989catastrophic, pmlr-v97-houlsby19a}, in particular for longer iterations of adaptation. To overcome these limitations, alternatives such as \textit{adapters}~\citep{rebuffi-redidual-adapters-2017, pmlr-v97-houlsby19a}, and \textit{sparse fine-tuning}~\citep{guo-etal-2021-parameter, ben-zaken-etal-2022-bitfit} have been introduced. These approaches, however, are still parameter- and time-inefficient, as they either add additional parameters or require complex training steps and/or models.

% meta-embeddings: concatenation~\citep{yin-schutze-2016-learning}, average~\citep{coates-bollegala-2018-frustratingly}, attention~\citep{kiela-etal-2018-dynamic}, adversarial feature learning~\citep{lange-etal-2021-fame}.

In this work, we propose \textbf{T}ask-\textbf{A}gnostic \textbf{D}omain \textbf{A}daptation for transformers (\TADA), a novel domain specialization framework. As depicted in Figure~\ref{fig:TADA}, it consists of two steps: (1) We conduct intermediate training of a pre-trained transformer-based language model (e.g., BERT) on the unlabeled domain-specific text corpora in order to inject domain knowledge into the transformer. Here, we \textit{fix} the parameter weights of the encoder while updating only the weights of the embeddings (i.e., embedding-based domain-adaptive pre-training). As a result, we obtain domain-specialized embeddings for each domain with the \textit{shared} encoder from the original PTLM without adding further parameters for domain adaptation. (2) The obtained domain-specialized embeddings along with the encoder can then be fine-tuned for downstream tasks in single- or multi-domain scenarios \citep{lange-etal-2021-share}, where the latter is conducted with meta-embeddings~\citep{coates-bollegala-2018-frustratingly, kiela-etal-2018-dynamic} and a novel meta-tokenization method for different tokenizers. 
\vspace{-0.6em}
\paragraph{Contributions.} We advance the field of domain specialization with the following contributions: \\
(i) We propose a modular, parameter-efficient, and task-agnostic domain adaptation method (\TADA) without introducing additional parameters for intermediate training of PTLMs. %This allows the domain-specialized embeddings to be portable to any downstream task for the respective target domain. 
(ii) We demonstrate the effectiveness of our specialization method on four heterogeneous downstream tasks -- dialog state tracking (DST), response retrieval (RR), named entity recognition (NER), and natural language inference (NLI) across 14 domains. 
%in both high- and low-resource settings. 
(iii) We propose modular domain specialization via meta-embeddings %~\citep{coates-bollegala-2018-frustratingly, kiela-etal-2018-dynamic}
and show the advantages %of embedding-based domain specialization
in multi-domain scenarios. 
(iv) We introduce the concept of meta-tokenization to combine sequences from different tokenizers in a single transformer model and perform the first study on this promising topic. (v) We release the code and resources for \TADA publicly.\footnote{\url{https://github.com/boschresearch/TADA}} % under an open-source license. %and the domain-specialized models on the huggingface model hub. 
%\vspace{-0.4em}
%\textcolor{cyan}{The contribution paragraph feels quite long. We can also move the code release part completely into a footnote now that we have many contributions}

\section{Methods for Domain Specialization}
\label{sec:methodology}
To inject domain-specific knowledge through domain-adaptive pre-training into PTLMs, these models are trained on unlabeled in-domain text corpora. For this, we introduce a novel \textit{embedding-based} intermediate training approach as an alternative to \textit{fully pre-training} and \textit{adapters} (\autoref{subsec:domain_specialization}), and further study the effects of domain-specific tokenization (\autoref{subsec:domain_specific_tokenization}). 
We then utilize multiple domain-specialized embeddings with our newly proposed meta-tokenizers and powerful meta-embeddings in multi-domain scenarios (\autoref{subsec:meta-embeddings} and \autoref{subsec:meta-tokenization}). 
%\vspace{-0.2em}
\subsection{Domain Specialization}
\label{subsec:domain_specialization}
Following successful work on \textit{intermediate pre-training} leveraging language modeling for domain-adaptation~\citep{gururangan-etal-2020-dont, hung-etal-2022-ds} and language-adaptation~\citep{glavas-etal-2020-xhate, hung-etal-2022-multi2woz}, %TODO: back in crv
we investigate the effects of training with masked language modeling (MLM) on domain-specific text corpora (e.g., clinical reports or academic publications). %The objective is to minimize the loss in predicting the masked tokens. 
For this, the MLM loss $L_{mlm}$ is commonly computed as the negative log-likelihood of the true token probability~\citep{devlin-etal-2019-bert, liu2019roberta}.
\begin{equation}
   L_{mlm}=-\sum_{m=1}^{M}\log P(t_m)\,,
\end{equation}

\normalsize
\noindent where $M$ is the total number of masked tokens in a given text and $P(t_m)$ is the predicted probability of the token $t_m$ over the vocabulary size. 

%\paragraph{Fully Fine-tuning.}
% problems of full MLMing, introduce variants and discuss more about it, and state our proposed approach, separate by paragraphs?
Fully pre-training the model requires adjusting all of the model's parameters, which can be undesirable due to time- and resource-inefficiency and can dramatically increase the risk of catastrophic forgetting of the previously acquired knowledge~\citep{mccloskey1989catastrophic, ansell-etal-2022-composable}. To alleviate these issues, we propose a parameter-efficient approach without adding additional parameters during intermediate domain-specialized adaptation: we freeze most of the PTLM parameters and only update the input embeddings weights of the first transformer layer (i.e., the parameters of the embeddings layer) during MLM. With this, the model can learn domain-specific input representations while preserving acquired knowledge in the frozen parameters. As shown in Figure~\ref{fig:TADA}, the encoder parameters are fixed during intermediate pre-training while only the embeddings layer parameters are updated. 

As a result, after intermediate MLM, multiple embeddings specialized for different domains are all applicable with the \textit{same} shared encoder. As these trained domain-specialized embeddings are easily \textit{portable} to any downstream task, we experiment with their combination in multi-domain scenarios via meta-embeddings methods~\citep{yin-schutze-2016-learning, kiela-etal-2018-dynamic}. We discuss this in more detail in Section \autoref{subsec:meta-embeddings}.  

\subsection{Domain-Specific Tokenization}
\label{subsec:domain_specific_tokenization}
Inspired by previous work on domain-specialized tokenizers and vocabularies for language model pre-training~\citep{beltagy-etal-2019-scibert, lee-etal-2019-biobert, yang2020finbert}, we study the domain adaptation of tokenizers for transformers and train domain-specialized variants with the standard WordPiece algorithm~\citep{schuster-etal-2012-wordpiece} analogously to the BERT tokenizer. As a result, the domain-specialized tokenizers cover more in-domain terms compared to the original PTLM tokenizers. In particular, this reduces the number of out-of-vocabulary tokens, i.e., words that have to be split into multiple subwords, whose embedding quality often does not match the quality of word-level representations~\citep{hedderich-etal-2021-survey}.

\subsection{Meta-Embeddings}
\label{subsec:meta-embeddings}
Given $n$ embeddings from different domains $D$, each domain would have an input representation $x_{Di} \in \mathbb{R}^{E}$, $1 \leq i \leq n$, where $n$ is the number of domains and $E$ is the dimension of the input embeddings. Here, we consider two variants: \textit{averaging}~\citep{coates-bollegala-2018-frustratingly} and \textit{attention-based} meta-embeddings~\citep{kiela-etal-2018-dynamic}.

Averaging merges all embeddings into one vector without training additional parameters by taking the unweighted average:
\begin{equation}
   e^{AVG}=\frac{1}{n}\sum_{i} x_{Di}\,,
\end{equation}
\normalsize
In addition, a weighted average with dynamic attention weights $\alpha_{Di}$ can be used. For this, the attention weights are computed as follows:
\begin{equation}
   \alpha_{Di}=\frac{exp(V \cdot tanh(W x_{Di}))}{\sum_{k=1}^{n}exp(V\cdot tanh(W x_{Dk}))}\,,
\end{equation}
\normalsize
\noindent with $W \in \mathbb{R}^{H \times E}$ and $V \in \mathbb{R}^{1 \times H}$ being parameters that are randomly initialized and learned during training and $H$ is the dimension of the attention vector which is a predefined hyperparameter. 

The domain embeddings $x_{Di}$ are then weighted using the learned attention weights $\alpha_{Di}$ into one representation vector:
\begin{equation}
   e^{ATT}=\sum_{i}\alpha_{Di}\cdot x_{Di}\,,
\end{equation}
%\vspace{-0.2em}
\normalsize
As \textit{Averaging} simply merges all information into one vector, it cannot focus on valuable domain knowledge in specific embeddings. 
In contrast, the \textit{attention-based} weighting allows for dynamic combinations of embeddings based on their importance depending on the current input token. 

\setlength{\tabcolsep}{2.6pt}
\begin{table*}[t]
\centering
\scriptsize{
\begin{tabular}{cccc}
\multicolumn{2}{c}{\multirow{2}{*}{\textbf{Domain Text:} Acetaminophen is an analgesic drug}}
   & \multicolumn{2}{l}{=> \textbf{TOK-1:} Ace \#ta \#mino \#phen is an anal \#gesic dr \#ug \hspace{0.25cm} (10 subwords)} \\
 & & \multicolumn{2}{l}{=> \textbf{TOK-2:} Aceta \#minophen is an anal \#gesic drug \hspace{0.80cm} (7 subwords)} \\ \ \\
 \toprule
 \multicolumn{2}{l}{\textbf{Aggregation:} \hspace{1.45cm} \textbf{\tokSpace}}
 %& \multicolumn{1}{|c}{\textbf{subword}} 
 & \multicolumn{1}{|c}{\textbf{\tokDynamic}} & \multicolumn{1}{|c}{\textbf{\tokTruncate}} \\ \cmidrule{1-1}\cmidrule(lr){2-2}\cmidrule(lr){3-3}\cmidrule(lr){4-4}
\textbf{TOK-1  } &  \multicolumn{1}{c}{[Ace \#ta \#mino \#phen] is an [anal \#gesic] [dr \#ug]} & \multicolumn{1}{c}{[Ace \#ta] [\#mino \#phen] is an anal \#gesic [dr \#ug]} & \multicolumn{1}{c}{[Ace] [\#mino] is an anal \#gesic [dr]} \\
\textbf{TOK-2  } &  \multicolumn{1}{c}{[Aceta \#minophen] is an [anal \#gesic] drug} & \multicolumn{1}{c}{Aceta \#minophen is an anal \#gesic drug} & \multicolumn{1}{c}{Aceta \#minophen is an anal \#gesic drug} \\
\bottomrule 
%\\
%& length of the original sequence & length of the shortest tokenizer sequence & length of the shortest tokenizer sequence \\
\end{tabular}%
}
%\vspace{-0.3em}
\caption{Examples of our proposed aggregation approaches for meta-tokenization: \textit{\tokSpace}, \textit{\tokDynamic}, \textit{\tokTruncate} for a given text and two different tokenizers (\textsc{TOK-1}, \textsc{TOK-2}). The bottom of the table shows the results after aggregation. $[a~b \dots z]$ denotes the average of all embedding vectors corresponding to subword tokens $a$, $b$, $\dots$, $z$.}
%\vspace{-0.4em}
\label{tab:example-aggregation}
\end{table*}
 
As shown in related works, these meta-embeddings approaches suffered from critical mismatch issues when combining embeddings of different sizes and input granularities (e.g., character- and word-level embeddings) that could be addressed by learning additional mappings to the same dimensions on word-level to force all the input embeddings towards a common input space~\citep{lange-etal-2021-fame}. 

Our proposed method prevents these issues by (a) keeping the input granularity fixed, which alleviates the need for learning additional mappings, and (b) locating all domain embeddings in the same %embedding 
space immediately after pre-training by freezing the subsequent transformer layers. We compare the results of two variants in Section~\autoref{sec:results}. More information on meta-embeddings can be found in the survey of~\citet{bollegala2022survey}.

\subsection{Meta-Tokenization for Meta-Embeddings}
\label{subsec:meta-tokenization}

%%Add dataset table
\setlength{\tabcolsep}{6.2pt}
\begin{table*}[t]
\centering
\scriptsize{
\begin{tabular}{lllrcc}
\toprule
\textbf{Task}                     & \textbf{Dataset}                                     & \textbf{Domain}     & \textbf{Background}        & \textbf{Train / Dev / Test}   & \textbf{License\dag}     \\ \midrule
\multirow{5}{*}{DST, RR} & \multirow{5}{*}{MultiWOZ 2.1~\citep{eric-etal-2020-multiwoz}}                   & Taxi       & 200 K & 1,654 / 207 / 195     & \multirow{5}{*}{MIT} \\
                         &                                             & Restaurant &  200 K                     & 3,813 / 438 / 437     &                      \\
                         &                                             & Hotel      &  200 K                     & 3,381 / 416 / 394     &                      \\
                         &                                             & Train      &  200 K                     & 3,103 / 484 / 494     &                      \\
                         &                                             & Attraction &  200 K                     & 2,717 / 401 / 395     &                      \\ \midrule
\multirow{5}{*}{NLI}     & \multirow{5}{*}{MNLI~\citep{williams-etal-2018-broad}}                       & Government & 46.0 K                 & 77,350 / 2,000 / 2,000     &  OANC   \\
                         &                                             & Travel     & 47.4 K                 & 77,350 / 2,000 / 2,000     &       OANC               \\
                         &                                             & Slate      & 214.8 K                & 77,306 / 2,000 / 2,000     &     OANC                 \\
                         &                                             & Telephone  & 234.6 K                & 83,348 / 2,000 / 2,000     &          OANC            \\
                         &                                             & Fiction    & 299.5 K                & 77,348 / 2,000 / 2,000     &        CC-BY-SA-3.0; CC-BY-3.0              \\\midrule
\multirow{5}{*}{NER}     & CoNLL~\citep{tjong-kim-sang-de-meulder-2003-introduction} & News       & 51.0 K                 & 14,987 / 3,466 / 3,684  &     DUA                 \\
& I2B2-CLIN~\citep{uzuner20112010}             & Clinical   & 299.9 K                & 13,052 / 3,263 / 27,625 &        DUA              \\
                         & SEC~\citep{salinas-alvarado-etal-2015-domain}         & Financial  & 4.8 K                  & 825 / 207 / 443      &              CC-BY-3.0        \\
                         & LITBANK~\citep{bamman-etal-2019-annotated}               & Fiction    & 299.5 K                & 5,548 / 1,388 / 2,973   &             CC-BY-4.0         \\
                         
                         & SOFC~\citep{friedrich-etal-2020-sofc}               & Science    & 300.1 K                & 489 / 123 / 263      &  CC-BY-4.0 \\
                         \bottomrule
\end{tabular}%
}
%\vspace{-0.3em}
\caption{Overview of the selected datasets for 4 tasks (DST, RR, NLI, NER) on 14 domains. For each domain, we report the number of collected in-domain texts for domain-adaptive pre-training, as well as the size and license of the downstream dataset. All selected datasets are applicable for \textit{commercial} usage. \dag License: Open American National Corpus (OANC), Direct Universal Access (DUA), Creative Commons Attribution Share-Alike (CC-BY-SA), Creative Commons Attribution International License (CC-BY).}
%\vspace{-1.5em}
\label{tab:data-analysis}
\end{table*}

To utilize our domain-adapted tokenizers in a single model with meta-embeddings, we have to align different output sequences generated by each tokenizer for the same input. This is not straightforward due to mismatches in subword token boundaries and sequence lengths. We thus introduce three different aggregation methods to perform the meta-tokenization: \\
(a) \tokSpace: We split the input sequence on 
\hyphenation{white-spaces}whitespaces into tokens and aggregate for each tokenizer all subword tokens corresponding to a particular token in the original sequence. \\
(b) \tokDynamic: The shortest sequence from all tokenizers is taken as a reference. Subwords from longer sequences are aggregated accordingly. This assumes %is based on the assumptions, 
that word-level knowledge is more useful than subword knowledge and that fewer word splitting is an indication of in-domain knowledge. \\
(c) \tokTruncate: This method is similar to the \tokDynamic aggregation, but it uses only the first subword for each token instead of computing the average when a token is split into more subwords. 

Once the token and subword boundaries are determined, we retrieve the subword embeddings from the embedding layer corresponding to the tokenizer and perform the aggregation if necessary, in our case averaging all subword embeddings. 
Examples for each method are shown in Table~\ref{tab:example-aggregation}. 
%\vspace{-0.4em}

\section{Experimental Setup}
This section introduces four downstream tasks with their respective datasets and evaluation metrics. We further provide details on our models, their hyperparameters, and the baseline systems. % for comparison. 
%\vspace{-0.4em}
\subsection{Tasks and Evaluation Measures}
We evaluate our domain-specialized models and baselines on four prominent downstream tasks: dialog state tracking (DST), response retrieval (RR), named entity recognition (NER), and natural language inference (NLI) with five domains per task. Table~\ref{tab:data-analysis} shows the statistics of all datasets. % for all tasks. 

%\paragraph{DST.}
\textbf{DST} is cast as a multi-classification dialog task. Given a dialog history (sequence of utterances) and a predefined ontology, the goal is to predict the output state, i.e., (domain, slot, value) tuples~\citep{wu-etal-2020-tod} like (\textit{restaurant}, \textit{pricerange}, \textit{expensive}). The standard joint goal accuracy is adopted as the evaluation measure: at each dialog turn, it compares the predicted dialog states against the annotated ground truth. The predicted state is considered accurate if and only if all the predicted slot values match exactly to the ground truth.

%\paragraph{RR.} 
\textbf{RR} is a ranking task, relevant for retrieval-based task-oriented dialog systems~\citep{henderson-etal-2019-training,wu-etal-2020-tod}. Given the dialog context, the model ranks $N$ dataset utterances, including the \textit{true response} to the context (i.e., the candidate set covers one \textit{true} response and $N-1$ \textit{false} responses). Following~\citet{henderson-etal-2019-training}, we report the recall at top rank given 99 randomly sampled false responses, denoted as $R_{100}@1$.

%\paragraph{NER.}
\textbf{NER} is a sequence tagging task, aiming to detect named entities within a sentence by classifying each token into the entity type from a predefined 
set of categories (e.g., \textsc{PERSON}, \textsc{ORGANIZATION}) including a neutral type (\textsc{O}) for non-entities. Following prior work~\citep{tjong-kim-sang-de-meulder-2003-introduction, nadeau2007survey}, we report the strict micro $F_1$ score. % as the evaluation metric.

%\paragraph{NLI.} 
\textbf{NLI} is a language understanding task testing the reasoning abilities of machine learning models beyond simple pattern recognition. The task is to determine if a \textit{hypothesis} logically follows the relationship from a \textit{premise}, inferred by \textsc{entailment} (true), \textsc{contradiction} (false), or \textsc{neutral} (undefined). Following \citet{williams-etal-2018-broad}, accuracy is reported as the evaluation measure.
%\vspace{-0.4em}
\subsection{Background Data for Specialization}
We take unlabeled background datasets from the original or related text sources to specialize our models with domain-adaptive pre-training (details are available in Appendix~\ref{appendix:domain-corpora}). For MLM training, we randomly sample up to 200K domain-specific sentences%contexts
\footnote{Except for four low-resource domains. For these, we randomly sample 44K (\textsc{Government, Travel, News}) and 4.5K (\textsc{Financial}) respectively.} and dynamically mask 15\% of the subword tokens following~\citet{liu2019roberta}.
%\vspace{-0.4em}
\subsection{Models and Baselines}
We experiment with the most widely used PTLM: BERT~\citep{devlin-etal-2019-bert} for NER and NLI. For DST and RR as dialog tasks, we experiment with BERT and TOD-BERT~\citep{wu-etal-2020-tod} following~\citet{hung-etal-2022-ds} for comparing general- and task-specific PTLMs.\footnote{We use the pre-trained models from HuggingFace: \texttt{bert-base-uncased} (NLI, NER) and \texttt{bert-base-cased}, \texttt{TODBERT/TOD-BERT-JNT-V1} (RR, DST).} We want to highlight that our proposed method can be easily applied to any existing PTLM. As baselines, we report the performance of the non-specialized variants and compare them against (a) full pre-training~\citep{gururangan-etal-2020-dont},  (b) adapter-based models~\citep{pmlr-v97-houlsby19a}, and (c) our domain-specialized PTLM variants trained with \TADA. 
%\vspace{-0.4em}
\subsection{Hyperparameters and Optimization}
\label{subsec:hyperparameters_and_optimization}
During %domain-adaptive 
MLM training, we fix the maximum sequence length to 256 (DST, RR) and 128 (NER, NLI) subwords and do lowercasing. We train for 30 epochs in batches of 32 instances and search for the optimal learning rate among the following values: $\{5\cdot 10^{-5}, 1\cdot 10^{-5}, 1\cdot 10^{-6}\}$. Early stopping is applied on the development set performance (patience:~3~epochs) and the cross-entropy loss is minimized using AdamW~\citep{loshchilov2018decoupled}. For DST and RR, we follow the hyperparameter setup from~\citet{hung-etal-2022-ds}. For NLI, we train for 3 epochs in batches of 32 instances. For NER, we train 10 epochs in batches of 8 instances. Both tasks use a fixed learning rate of $5\cdot 10^{-5}$.

% \begin{table*}[t]
% \centering
% \scriptsize{
% \begin{tabular}{lllllllll}
% \toprule
% \multicolumn{3}{c}{\textbf{MultiWOZ}}               & \multicolumn{3}{c}{\textbf{MNLI}}         & \multicolumn{3}{c}{\textbf{NER}}                \\\cmidrule(lr){1-3}\cmidrule(lr){4-6}\cmidrule(lr){7-9}
% Domain     & Specialization        & Downstream     & Domain     & Specialization & Downstream  & Domain    & Specialization & Downstream         \\
% Taxi       & \multirow{5}{*}{200K} & 1654, 207, 195 & Government & 46.0K          & 77350, 2000 & Financial & 4.8K           & 825, 207, 443      \\
% Restaurant &                       & 3813, 438, 437 & Telephone  & 234.6K         & 83348, 2000 & News      & 51.0K          & 14987, 3466, 3684  \\
% Hotel      &                       & 3381, 416, 394 & Fiction    & 299.5K         & 77348, 2000 & Fiction   & 299.5K         & 5548, 1388, 2973   \\
% Train      &                       & 3103, 484, 494 & Slate      & 214.8K         & 77306, 2000 & Clinical  & 299.9K         & 13052, 3263, 27625 \\
% Attraction &                       & 2717, 401, 395 & Travel     & 47.4K          & 77350, 2000 & Science   & 300.1K         & 489, 123, 263     \\
% \bottomrule
% \end{tabular}%
% }
% \caption{}
% \label{tab:all_data}
% \end{table*}

\section{Evaluation Results}
\label{sec:results}
For each downstream task, we first conduct experiments in a single-domain scenario, i.e., training and testing on data from the same domain, to show the advantages of our proposed approach of task-agnostic domain-adaptive embedding-based pre-training and tokenizers (\autoref{subsec:singledomain}). We further consider the combination of domain-specialized embeddings with meta-embeddings variants~\citep{coates-bollegala-2018-frustratingly, kiela-etal-2018-dynamic} in a multi-domain scenario, where we jointly train on data from all domains of the respective task (\autoref{subsec:multidomain}).

\subsection{Single-Domain Evaluation}
\label{subsec:singledomain}
\setlength{\tabcolsep}{2.1pt}
\begin{table*}[th]
\centering
\scriptsize{
\begin{tabular}{l ccccc g ccccc g}
\toprule
& \multicolumn{6}{c}{\textbf{DST}}     & \multicolumn{6}{c}{\textbf{RR}}          \\ 
\textbf{Model} & \textbf{Taxi} & \textbf{Restaurant} & \textbf{Hotel} & \textbf{Train} & \textbf{Attraction} & \textbf{Avg.} &  \textbf{Taxi} & \textbf{Restaurant} & \textbf{Hotel} & \textbf{Train} & \textbf{Attraction} & \textbf{Avg.}\\ \cmidrule(lr){2-7}\cmidrule(lr){8-13}
BERT                 & 23.87 & 35.44      & 30.18 & 41.93 & 29.77      & 32.24 & 23.25 & 37.61      & 38.97 & 44.53 & 48.47      & 38.57 \\
TOD-BERT             & 30.45 & 43.58      & 36.20 & 48.79 & 42.70      & 40.34 & 45.68 & 57.43      & 53.84 & 60.66 & 60.26      & 55.57 \\\midrule
BERT (\mlmFull)             & 23.74 & 37.09      & 32.77 & 40.96 & 36.66      & 34.24 &   31.37    &        53.08    &     45.41  &   51.66    &       52.23     &     46.75   \\
TOD-BERT (\mlmFull)         & 29.94 & 43.14      & 36.11 & 47.61 & 41.54      & 39.67 & 41.77 & 55.27      & 50.60  & 55.17 & 54.62      & 51.49 \\ 
BERT (\mlmAdapt)             & 22.52 & 40.49 & 31.90 & 42.17 & 35.05 & 34.43 & 32.84 & 44.01 & 39.15 & 38.43 & 45.05 & 39.90   \\
TOD-BERT (\mlmAdapt)         & 32.06 & 44.06 & 36.74 & \textbf{48.84} & 43.50 & \textbf{41.04}  & 49.08 & 58.18 & 55.55 & 59.46 & 60.26 & 56.51 \\ \midrule
%\textit{Embedding-based: TADA} & & & & & & \multicolumn{1}{c}{} &  & & & & & \multicolumn{1}{c}{}\\
BERT (\mlmEmb)             & 22.39 & 31.26      & 25.75 & 41.00 & 34.02      & 30.88 &   40.89 & 54.24      & 47.30 & 52.18 & 56.50     &    50.22   \\
TOD-BERT (\mlmEmb)         & 32.00 & 43.47      & 36.67 & 47.34 & 42.80      & 40.46 & 47.08 & 57.71      & 55.65  & 60.72 & 60.39 & 56.31 \\ 
TOD-BERT (\mlmEmbTokS)             & \textbf{33.03} & 41.14      & 36.77 & 47.50 & 40.77      & 39.84 &   50.41 & 58.97      & 56.48 & \textbf{62.63} & 59.56     &    57.61   \\
TOD-BERT (\mlmEmbTokX) & 32.55 & \textbf{44.60}      & \textbf{36.92} & 47.27 & \textbf{43.58}      & 40.98 & \textbf{50.77} & \textbf{60.40}      & \textbf{56.87}  & 62.11 & \textbf{60.89} & \textbf{58.21} \\ 
%&&&&&&&&&&&& \\
  \midrule
%\end{tabular}%
%}
%\label{tab:eval_result_single}
%%\vspace{-0.5em}
%\end{table*}
%\setlength{\tabcolsep}{4.2pt}
%\begin{table*}[!ht]
%\centering
%\scriptsize{
%\begin{tabular}{l ccccc g ccccc g}
\midrule
 & \multicolumn{6}{c}{\textbf{NLI}}    & \multicolumn{6}{c}{\textbf{NER}}    \\ 
\textbf{Model} & \textbf{Government} & \textbf{Telephone} & \textbf{Fiction} & \textbf{Slate} & \textbf{Travel} & \textbf{Avg.} & \textbf{Financial} & \textbf{Fiction} & \textbf{News} & \textbf{Clinical} & \textbf{Science} &  \textbf{Avg.}\\ \cmidrule(lr){2-7}\cmidrule(lr){8-13}
BERT                  & 79.07 & 78.18 & 76.63 & 73.40 & 77.33 & 76.92 &\textbf{90.56} & 72.09  & 90.04 & 85.91 & 78.23 & 83.44\\\midrule
BERT (\mlmFull)              & 80.82 & \textbf{81.43} & 76.43 & 71.97 & 77.78 & 77.69 & 90.53 & \textbf{72.33}  & 90.62 & \textbf{86.18} & 78.19 & \textbf{83.57}\\
BERT (\mlmAdapt)                 & 75.58 & 73.70 & 72.33 & 67.11 & 72.42 & 72.23 & 76.62 & 63.82 & 89.17 & 80.64 & 61.65 & 74.38\\\midrule
BERT (\mlmEmb)            & 80.77 & 80.42 & \textbf{79.27} & \textbf{73.50} & \textbf{77.94} & \textbf{78.38} & 90.38  &  71.79  & \textbf{90.67} & 85.82 & 78.82 & 83.50 \\
BERT (\mlmEmbTokS)          & 80.57 & 79.15 & 78.51 & 72.94 & 77.28 & 77.69 & 87.49 & 69.90 & 89.55 & 85.53 & \textbf{79.39} & 82.37\\
BERT (\mlmEmbTokX)           & \textbf{81.08} & 80.16 & 78.97 & 73.15 &  77.68 & 78.21& 89.27 & 69.77 & 89.21 & 85.31 & 77.33 & 82.18\\
  \bottomrule
\end{tabular}%
}

\caption{Results of our single-domain models with domain-specialized embeddings and tokenizers on four tasks.}
    %\vspace{-1.6em}
%\caption{Results of single domain evaluation on two downstream tasks: Dialog State Tracking (DST) and Response Retrieval (RR) with joint goal accuracy (\%) as the metric for DST and $\textsc{R}_{100}@1$ (\%) for RR. The baseline results are inherited from~\citet{hung-etal-2022-ds}. \textcolor{cyan}{TODO: Restructure table 2 and 3 to mach style of table 4}}
%\caption{Results of single domain results on NLI and NER task with accuracy (\%) and F1 (\%) as the evaluation metric.}
\label{tab:eval_result_mnli_ner_single}
%\vspace{-0.5em}
\end{table*}
We report downstream performance for the single-domain scenario in Table~\ref{tab:eval_result_mnli_ner_single}, with each subtable being segmented into three parts: (1) at the top, we show baseline results (BERT, TOD-BERT) without any domain specialization; (2) in the middle, we show results of domain-specialized PTLMs via full domain-adaptive training and the adapter-based approach; (3) the bottom of the table contains results of our proposed approach specializing only the embeddings and the domain-specific tokenization. 

In both DST and RR, TOD-BERT outperforms BERT due to its training for conversational  knowledge. By further domain-adaptive pre-training with full MLM training (\mlmFull), TOD-BERT's performance decreases (i.e., -4\% for RR and -0.8\% for DST compared to TOD-BERT). It is argued that full MLM domain specialization has negative interference: while TOD-BERT is being trained on domain data during intermediate pre-training, the model is forgetting the conversational knowledge obtained during the initial dialogic pre-training stage~\citep{wu-etal-2020-tod}. The hypothesis is further supported by the observations for the adapter-based method which gains slight performance increases. %This is aligned with the previous findings of~\citet{qiu-etal-2021-different} and~\citet{hung-etal-2022-ds}. 

Our proposed embedding-based domain-adaptation (\mlmEmb) yields similar performance gains as specialization with adapters for TOD-BERT on average. Inspired by previous work on domain-specialized subtokens for language model pre-training~\citep{beltagy-etal-2019-scibert, yang2020finbert}, we additionally train domain-specific tokenizers (\mlmEmbTok) with the %standard 
WordPiece algorithm~\citep{schuster-etal-2012-wordpiece}. %Here, 
The training corpora %for domain-specialized tokenizers 
are either obtained from only %the 
background corpora (\textbf{S}) or from the combination of background and training set of each domain (\textbf{X}). Further, our domain-specialized tokenizers coupled with the embedding-based domain-adaptive pre-training exhibit similar average performance for DST and outperform the state-of-the-art adapters and all other methods for RR. 

Similar findings are observed for NLI and NER. %These are reported in Table~\ref{tab:eval_result_mnli_ner_single}. % with comparisons between non-domain-specialized method and domain-adaptive pre-training approaches (MLM-FULL vs MLM-ADAPT vs MLM-EMB). 
\mlmEmb compared to \mlmFull results in +0.7\% performance gains in NLI and reaches similar average gains in NER. Especially for NLI, the domain-specialized tokenizers (\mlmEmbTok) are beneficial in combination with our domain-specialized embeddings, while having considerably fewer trainable parameters. 
Given that \TADA %our proposed approach 
is substantially more efficient and parameter-free (i.e., without adding extra parameters), % for each layer), % compared to adapter-based)
this promises more sustainable domain-adaptive pre-training.

\subsection{Multi-Domain Evaluation}
\label{subsec:multidomain}

\setlength{\tabcolsep}{1.2pt}
\begin{table*}[th]
\centering
\scriptsize{
\begin{tabular}{l ccccc g ccccc g}
\toprule
& \multicolumn{6}{c}{\textbf{DST}}     & \multicolumn{6}{c}{\textbf{RR}}          \\ 
\textbf{Model} & \textbf{Taxi} & \textbf{Restaurant} & \textbf{Hotel} & \textbf{Train} & \textbf{Attraction} & \textbf{Avg.} &  \textbf{Taxi} & \textbf{Restaurant} & \textbf{Hotel} & \textbf{Train} & \textbf{Attraction} & \textbf{Avg.}\\ \cmidrule(lr){2-7}\cmidrule(lr){8-13}
BERT                 & 29.10 & 39.92      & 36.67 & 47.63 & 42.32      & 39.13 & 44.87 & 51.98      & 49.11 & 50.15 & 54.81      & 50.18 \\
TOD-BERT             & 34.65 & 44.24      & 39.54 & 51.66 & 44.24      & 42.87 & 50.99 & 61.53      & 56.09 & 58.94 & 62.76      & 58.06 \\
BERT (\mlmFull)             & 31.94 & 42.16 & 38.48 & 45.37 & 41.48 & 39.89 & 49.59 & 55.76& 54.66 & 55.59 & 59.85& 55.09  \\
TOD-BERT (\mlmFull)         & 32.26 & 45.70 & 39.51 & 51.31 & 45.92 & 42.94 & 53.51 & 64.44 & 59.22  &  62.14 & \textbf{66.49} & 61.16 \\\midrule
(AVG) TOD-BERT (EMB+\mlmEmbs)             & \textbf{37.65} & 46.06      & 39.61 & 51.95 & 46.95     & 44.44 &   52.84 & 62.56      & 58.54 & 60.79 & 64.87     &    59.92  \\
(ATT) TOD-BERT (EMB+\mlmEmbs)          & 35.13 & 46.86      & 40.73 & 51.10 & 44.76  & 43.72 & 53.06 & 63.18      & 56.94  & 60.45 & 64.13 & 59.55 \\
(AVG) TOD-BERT (\mlmEmbs)          & 35.42 & 46.71 & 40.82 & \textbf{52.34}& 47.30 & 44.52 & \textbf{55.20} & \textbf{64.58} & \textbf{60.39} & \textbf{62.84} & 66.11 & \textbf{61.82} \\
(ATT) TOD-BERT (\mlmEmbs)          & 37.35 &  \textbf{46.98}     & \textbf{41.32}  & 51.92  & \textbf{47.88}  & \textbf{45.09} & 53.73 & 64.00     & 59.89 & 61.54 & 65.05 & 60.84 \\\midrule \midrule
& \multicolumn{6}{c}{\textbf{NLI}}    & \multicolumn{6}{c}{\textbf{NER}}    \\ 
\textbf{Model} & \textbf{Government} & \textbf{Telephone} & \textbf{Fiction} & \textbf{Slate} & \textbf{Travel} & \textbf{Avg.} & \textbf{Financial} & \textbf{Fiction} & \textbf{News} & \textbf{Clinical} & \textbf{Science} &  \textbf{Avg.}\\ \cmidrule(lr){2-7}\cmidrule(lr){8-13}
BERT                  & 82.88	&\textbf{82.10}&	80.69&	76.01&	80.11	&80.36& 87.68	& 69.11 &	89.96&	\textbf{85.76}&	76.14	&81.73  \\
BERT (\mlmFull)                 & 83.29 & 81.79 & 81.11	&	76.32 &	79.66	&  80.43 & 88.71 	& \textbf{69.92} &	89.69 &	 85.61 &	80.03	& 82.79 \\\midrule
(AVG) BERT (\mlmEmbs)         &\textbf{83.80}&	80.87&	81.70&	\textbf{77.60}&	\textbf{81.30}&	\textbf{81.05} & 87.72	&68.78	&90.16&	85.68&	78.22&	82.11 \\
(ATT) BERT (\mlmEmbs)            &83.50&	81.64	&\textbf{81.74}	&76.68	&80.36& 80.78& \textbf{88.89}&	69.05&	\textbf{90.56}&	85.43&	\textbf{80.55}&	\textbf{82.90}\\
  \bottomrule
\end{tabular}%
}
%\vspace{-0.6em}
\caption{Results of our multi-domain models leveraging meta-embeddings on four downstream tasks.}
\label{tab:eval_result_multi_tasks}
%\vspace{-0.7em}
\end{table*}

In practice, %real-world scenarios, 
a single model must be able to handle multiple domains because the %concurrent 
deployment of multiple %single-domain 
models may not be feasible. % due to limited computational resources. 
To simulate a multi-domain setting, we utilize the domain-specialized embeddings from each domain (\autoref{subsec:singledomain}) and combine them %for experiments
with meta-embeddings (\autoref{subsec:meta-embeddings}). 

To train a single model for each task applicable to all domains, we concatenate the training sets of all domains for each task. As baselines for DST and RR, we report the performance of BERT and TOD-BERT and a version fine-tuned on the concatenated multi-domain training sets (\mlmFull). We test the effect of multi-domain specialization in two variants: \textit{averaging} (AVG) and \textit{attention-based} (ATT) meta-embeddings. We conduct experiments to check whether including general-purpose embeddings from TOD-BERT (EMB+\mlmEmbs) is beneficial compared to the one without (\mlmEmbs). The results in Table~\ref{tab:eval_result_multi_tasks} show that combining domain-specialized embeddings outperforms TOD-BERT in both tasks.
In particular, averaging meta-embeddings performs better in RR while attention-based ones work better in DST by 3.8\% and 2.2\% compared to TOD-BERT, respectively. It is further suggested that combining only domain-specialized embeddings (i.e., without adding general-purpose embeddings) % from TOD-BERT) 
works better for both meta-embeddings variants. 

These findings are confirmed by NLI and NER experiments. The meta-embeddings applied in our multi-domain scenarios outperform BERT by 0.7 points for NLI and 1.2 points for NER, respectively. An encouraging finding is that two domains (\textsc{Financial, Science}) with the smallest number of training resources benefit the most compared to the other domains in the NER task. Such few-shot settings are further investigated in \autoref{subsec:few-shot}.  

Overall, we find that the meta-embeddings provide a simple yet effective way to combine %information from 
several domain-specialized embeddings, alleviating the need of deploying multiple %single-domain 
models. 
%\vspace{-0.2em}

\section{Analysis}
%\vspace{-0.5em}
To more precisely analyze the advantages of our proposed embedding-based domain-adaptive pre-training methods and tokenizers, %in high- and low-resource settings, 
we study the following: few-shot transfer capability (\autoref{subsec:few-shot}), the effect of domain-specialized tokenizers on subword tokens (\autoref{subsec:domain-tokens}), and the combinations of multiple domain-specialized tokenizers with meta-tokenizers in multi-domain scenarios (\autoref{subsec:meta-tokenizers-exp}).

%\vspace{-0.3em}
\subsection{Few-Shot Learning}
\label{subsec:few-shot}

\newcommand{\sd}[1]{\mathsize{$\pm$}\srcsize{#1}}
%#scriptscriptstyle\mathrel{\mathsmaller

\label{appendix:fs}
\setlength{\tabcolsep}{1.6pt}
\begin{table*}[h]
\centering
\scriptsize{
\begin{tabular}{cl ccccc ccccc gg}
\toprule
\multicolumn{1}{l}{} &  & \multicolumn{2}{c}{\textbf{Government}} & \multicolumn{2}{c}{\textbf{Telephone}} & \multicolumn{2}{c}{\textbf{Fiction}} & \multicolumn{2}{c}{\textbf{Slate}} & \multicolumn{2}{c}{\textbf{Travel}} & \multicolumn{2}{g}{\textbf{Avg.}} \\\cmidrule(lr){3-4}\cmidrule(lr){5-6}\cmidrule(lr){7-8}\cmidrule(lr){9-10}\cmidrule(lr){11-12}\cmidrule(lr){13-14}
\multicolumn{1}{l}{} & \textbf{\textbf{Model}} & \multicolumn{1}{c}{\textbf{1\%}} & \multicolumn{1}{c}{\textbf{20\%}} & \multicolumn{1}{c}{\textbf{1\%}} & \multicolumn{1}{c}{\textbf{20\%}} & \multicolumn{1}{c}{\textbf{1\%}} & \multicolumn{1}{c}{\textbf{20\%}} & \multicolumn{1}{c}{\textbf{1\%}} & \multicolumn{1}{c}{\textbf{20\%}} & \multicolumn{1}{c}{\textbf{1\%}} & \multicolumn{1}{c}{\textbf{20\%}} & \multicolumn{1}{g}{\textbf{1\%}} & \multicolumn{1}{g}{\textbf{20\%}} \\\midrule
\multirow{5}{*}{\textbf{SD}} & BERT & 57.62\sd{5.4} & 75.21\sd{.4} & 49.20\sd{1.9} & 74.45\sd{.3} & 43.76\sd{2.2} & 72.90\sd{.3} & 
                                      46.70\sd{2.1} & 67.71\sd{.5} & 54.05\sd{4.0} & 71.55\sd{.4} & 50.27\sd{2.4} & 72.36\sd{.1} \\
                              & BERT (\mlmFull) & \textbf{61.92}\sd{1.8} & 76.07\sd{.7} & \textbf{54.53}\sd{1.6} & 75.07\sd{.7}  & 49.32\sd{1.4} & \textbf{73.21}\sd{.6} &
                                                   45.81\sd{0.7}         & 67.26\sd{.6} & 56.56\sd{3.5}           & 72.50\sd{.4} & 53.63\sd{0.5} & 72.82\sd{.4} \\
                              & BERT (\mlmAdapt) & 42.88\sd{1.8} & 67.93\sd{.2} & 41.27\sd{1.1} & 65.80\sd{.2} & 38.12\sd{1.7} & 59.53\sd{.4} & 
                                                   38.91\sd{2.1} & 54.71\sd{.7} & 40.74\sd{2.8} & 65.89\sd{.6} & 40.38\sd{1.5} & 62.78\sd{.7} \\\cmidrule(lr){2-14}
                              & BERT (\mlmEmb) & 61.66\sd{1.0} & \textbf{76.61}\sd{.3} & 49.86\sd{0.8} & \textbf{75.33}\sd{.3} & 48.35\sd{4.1} & 72.22\sd{.6} & 
                                                \textbf{49.10}\sd{2.5} & \textbf{68.26}\sd{.3} & \textbf{60.27}\sd{1.6} & \textbf{72.73}\sd{.6} & \textbf{53.85}\sd{1.7} & \textbf{73.03}\sd{.1} \\ 
                              & BERT (\mlmEmbTokX) & 61.27\sd{1.8} & 75.75\sd{.5} & 49.20\sd{5.5} & 74.11\sd{.1} & \textbf{49.74}\sd{0.8} & 72.26\sd{.8} & 
                                                      \textbf{49.10}\sd{1.9} & 66.51\sd{.8} & 58.99\sd{2.3} & 72.15\sd{.8} & 53.66\sd{2.0} & 72.16\sd{.1} \\ \midrule\midrule
\multirow{3}{*}{\textbf{MD}} & BERT                 & 69.56\sd{3.2} & 79.49\sd{.7} & 64.80\sd{2.0} & 77.72\sd{.2} & 61.53\sd{2.5} & 76.84\sd{.7}
                                                     & 61.43\sd{2.0} & 72.64\sd{.4} & 66.40\sd{2.9} & \textbf{76.42}\sd{.5} & 64.74\sd{1.8} & 76.62\sd{.2} \\\cmidrule(lr){2-14}
                             & (AVG) BERT (\mlmEmbs) & 70.13\sd{1.3} & \textbf{80.00}\sd{.2} & 64.39\sd{1.3} & 78.28\sd{.2} & \textbf{62.24}\sd{1.7} & 76.94\sd{.4} & 
                                                      \textbf{62.61}\sd{1.6} & 71.61\sd{.3} & \textbf{66.45}\sd{1.4} & 76.21\sd{.4} & 65.16\sd{1.3} & 76.61\sd{.1}  \\
                             & (ATT) BERT (\mlmEmbs) &\textbf{71.21}\sd{1.1} & 79.90\sd{.3} & \textbf{65.56}\sd{1.4} & \textbf{78.48}\sd{.1} & 61.33\sd{1.3} & \textbf{77.34}\sd{.3} & 
                                                      61.99\sd{1.3} & \textbf{72.69}\sd{.4} & 66.24\sd{1.7} & 76.32\sd{.5} & \textbf{65.27}\sd{1.6} & \textbf{76.95}\sd{.2} \\
 \bottomrule
\end{tabular}%
}
%\vspace{-0.4em}
\caption{Few-shot learning results on NLI task for 1\% and 20\% of the training data size in single-domain (SD) and multi-domain (MD) scenarios. We report mean and standard deviation of 3 runs with different random seeds. %, reported in accuracy (\%). The results are reported with the mean and standard deviation with 3 runs based on different random seeds. 
}
%\vspace{-1.2em}
\label{tab:mnli_fs}
\end{table*}

We report few-shot experiments in Table~\ref{tab:mnli_fs} using 1\% and 20\% of the training data %($\sim$770 and 16 K instances per domain) 
%in both single- and multi-domain scenarios 
for NLI.
We run three experiments with different random seeds to reduce variance and report the mean and standard deviation for these limited data scenarios.
\mlmEmb on average outperforms \mlmFull by 1\% in the single-domain scenario, especially for \textsc{Slate} and \textsc{Travel} domains with the largest improvements (i.e., 3.3\% and 2.7\%, respectively). In contrast, the adapter-based models (\mlmAdapt) perform worse in this few-shot setting. This demonstrates the negative interference (-10\%) caused by the additional parameters that cannot be properly trained given the scarcity of task data for fine-tuning. In multi-domain settings, attention-based meta-embeddings on average surpass the standard BERT model in both few-shot setups. Overall, these findings demonstrate the strength of our proposed embedding-based domain-adaptive pre-training in limited data scenarios.
%\vspace{-0.8em}

\setlength{\tabcolsep}{2.5pt}
\begin{table}[t]
\centering
\scriptsize{
\begin{tabular}{l ccccc gg}
\toprule
& \multicolumn{7}{c}{\textbf{Dialog State Tracking} and \textbf{Response Retrieval}}          \\ 
\textbf{Model} & \textbf{Taxi} & \textbf{Restaur.} & \textbf{Hotel} & \textbf{Train} & \textbf{Attract.} & \textbf{Avg.} & \textbf{Diff.} \\ 
\cmidrule(lr){2-6}\cmidrule{7-8}
\tokO & 856 & 1597 & 1530 & 1659 & 1310 & 1390.4 & -\\
\tokS & 715 & 1338 & 951  & 951  & 946  & 1048.2 & -24.6\%\\
\tokX & 465 & 959  & 753  & 753  & 740  & 798.4 & -42.6\%\\
\midrule
& \multicolumn{7}{c}{\textbf{Natural Language Inference}}          \\ 
\textbf{Model} & \textbf{Govern.} & \textbf{Tele.} & \textbf{Fiction} & \textbf{Slate} & \textbf{Travel} & \textbf{Avg.} &  \textbf{Diff.}\\
\cmidrule(lr){2-8}
\tokO & 4095 & 4221 & 3379 & 5094 & 5883 & 4534.3 & -\\
\tokS & 1874 & 3517 & 3568 & 3597 & 3685 & 3248.2 & -28.4\%\\
\tokX & 1873 & 3522 & 2426 & 3683 & 3984 & 3097.6 & -31.7\%\\
\midrule
& \multicolumn{7}{c}{\textbf{Named Entity Recognition}}          \\ 
\textbf{Model} & \textbf{Financ.} & \textbf{Fiction} & \textbf{News} & \textbf{Clinical} & \textbf{Science} &  \textbf{Avg.} & \textbf{Diff.} \\
\cmidrule(lr){2-8}
\tokO & 397 & 1930 & 6357 & 5121 & 832 & 2927.4 & -\\
\tokS & 695 & 1958 & 8526 & 3744 & 653 & 3115.2 & +6.4\%\\
\tokX & 600 & 1822 & 5818 & 2939 & 463 & 2328.4 & -20.5\%\\
\bottomrule
\end{tabular}%
}
%\vspace{-0.4em}
\caption{The number of words that have to be split into multiple tokens (>= subwords) for different tokenizers. }
%\vspace{-1.0em}
\label{tab:subword-splits-tokenizers}
\end{table}
%\vspace{-0.3em}
\subsection{Domain-Specific Tokenizers}
\label{subsec:domain-tokens}
To study whether domain-specialized tokenizers better represent the target domain, we select the development 
sets and count the number of words that are %have to be 
split into multiple tokens for each tokenizer. The assumption is that the domain-specialized tokenizers allow for word-level segmentation, and thus, word-level embeddings, instead of fallbacks to lower-quality embeddings from multiple subword tokens. %Therefore, we check if the number of words required to be split %into multiple subtokens
%is less for the domain-specific tokenizers compared to the original tokenizers.

We compare three different tokenizers for each setting: (a) \tokO: original tokenizer from PTLMs without domain specialization; (b) \tokS: domain-specialized tokenizer trained on the in-domain background corpus; (c) \tokX: domain-specialized tokenizer trained on the concatenated in-domain background corpus plus the training set. % of each selected domain.

Table~\ref{tab:subword-splits-tokenizers} shows the results on all four tasks averaged across domains. It is evident that \tokX compared to \tokO in general significantly reduces the number of tokens split into multiple subwords (-42.6\% in DST, RR; -31.7\% in NLI; -20.5\% in NER). This indicates that the domain-specialized tokenizers cover more tokens on the word-level, and thus, convey more domain-specific information. For domains with smaller background datasets, e.g., \textsc{Financial} and \textsc{News}, the tokenizers are not able to leverage more word-level information. For example, \tokS that was trained on the background data performs worse in these domains, as the background data is too small and the models overfit on background data coming from a similar, but not equal source. Including the training corpora helps to avoid overfitting and/or shift the tokenizers towards the dataset word distribution, as \tokX improves for both domains over \texttt{TOK-S}. The finding is well-aligned with the results in Table~\ref{tab:eval_result_mnli_ner_single} (see \autoref{subsec:singledomain}) and supports our hypothesis that word-level tokenization is beneficial.

%\vspace{-0.6em}
\subsection{Study on Meta-Tokenizers}
\label{subsec:meta-tokenizers-exp}
\setlength{\tabcolsep}{3.1pt}
\begin{table}[t]
\centering
\scriptsize{
\begin{tabular}{l cccc}
\toprule
\textbf{Model} & \textbf{DST} & \textbf{RR} & \textbf{NLI} & \textbf{NER} \\ \cmidrule(lr){2-5}
(AVG) BERT$\ddag$ (\mlmEmbs)               & 44.52 & \textbf{61.82} & \textbf{81.05} & 82.11 \\
(ATT) BERT$\ddag$ (\mlmEmbs)               & \textbf{45.09} & 60.84 & 80.78 & \textbf{82.90} \\\midrule
(AVG) BERT$\ddag$ (\mlmEmbsTokX) dyn             &  42.16 & \underline{59.87} & 79.10 & 70.73  \\
(AVG) BERT$\ddag$ (\mlmEmbsTokX) space         & 41.57 & 58.54 & 79.51 & 70.63  \\
(AVG) BERT$\ddag$ (\mlmEmbsTokX) trun             & 40.26 & 58.07 & 79.47 & 66.66  \\
(ATT) BERT$\ddag$ (\mlmEmbsTokX) dyn         & \underline{42.73} & 59.22 & 79.32 & 
 \underline{70.83}\\
(ATT) BERT$\ddag$ (\mlmEmbsTokX) space             & 41.45 & 58.95 & \underline{79.93} &  70.71 \\
(ATT) BERT$\ddag$ (\mlmEmbsTokX) trun         & 40.82 & 59.09 & 79.67 & 68.41 \\

  \bottomrule
\end{tabular}%
}
%\vspace{-0.4em}
\caption{Results of meta-tokenizers in multi-domain experiments with meta-embeddings. Here \textbf{bold} indicates the best performance and \underline{underline} indicates the best-performing meta-tokenization aggregation method. $\ddag$BERT variants: TOD-BERT (DST, RR) and BERT (NLI, NER).
}
\label{tab:eval_result_metatok}
%\vspace{-1.5em}
\end{table}

%\vspace{-0.2em}
In Section \autoref{subsec:multidomain}, we experiment with %the combination of 
multiple domain-specialized embeddings inside meta-embeddings. These embeddings are, however, based on the original tokenizers and not on the domain-specialized ones. While the latter are considered to contain more domain knowledge %for %regarding more 
%word-level tokens (\autoref{subsec:domain-tokens}) 
and achieve better downstream single-domain %task 
performance (\autoref{subsec:singledomain}), it is not straightforward to combine tokenized output by different tokenizers for the same input due to mismatches in subword %token 
boundaries and sequence lengths. 

Therefore, we further conduct experiments with meta-tokenizers in the meta-embeddings setup following \autoref{subsec:meta-tokenization}. 
We compare the best multi-domain models with our proposed %tokenizer 
aggregation approaches. The averaged results across domains %per task 
are shown in Table~\ref{tab:eval_result_metatok} (per-domain results are available in Appendix~\ref{appendix:md-mt}). Overall, it is observed that the \tokSpace and \tokDynamic approaches work better than \tokTruncate. However, there is still a performance gap between using multiple embeddings sharing the same sequence from the original tokenizer compared to the domain-specialized tokenizers. Nonetheless, this study shows the general applicability of meta-tokenizers in transformers and suggests future work toward %effectively 
leveraging the domain-specialized tokenizers in meta-embeddings.
%\vspace{-0.5em}

\section{Related Work}
%\vspace{-0.4em}

\paragraph{Domain Adaptation.}
%\textcolor{blue}{importance of domain adaptation, list papers (CC) for both single and multi-domain related PTLMs}
Domain adaptation is a type of transfer learning that aims to enable the trained model to be generalized into a specific domain of interest~\citep{farahani2021brief}. 
Recent studies have focused on neural unsupervised or self-supervised domain adaptation leveraging PTLMs~\citep{ramponi-plank-2020-neural}, which do not rely on large-scale labeled target domain data to acquire domain-specific knowledge.~\citet{gururangan-etal-2020-dont} proposed domain-adaptive intermediate pre-training, continually training PTLM on MLM with domain-relevant unlabeled data, leading to improvements in downstream tasks in both high- and low-resource setups. The proposed approach has been applied to multiple tasks~\citep{glavas-etal-2020-xhate, lewis-etal-2020-pretrained} across languages~\citep{hung-etal-2023-demographic, wang2023nlnde}, however, requires \textit{fully} pre-training (i.e., update all PTLM parameters) during domain adaptation, which can potentially result in catastrophic forgetting and negative interference~\citep{pmlr-v97-houlsby19a, he-etal-2021-effectiveness}. 

\paragraph{Parameter-Efficient Training.}
Parameter-efficient methods for domain adaptation alleviate these problems. They have shown robust performance in low-resource and few-shot scenarios~\citep{fu2022effectiveness}, where only a small portion of parameters are trained while the majority of parameters are frozen and shared across tasks. 
These lightweight alternatives are shown to be more stable than their corresponding fully fine-tuned counterparts and perform \emph{on par with} or better than expensive fully pre-training setups, including \textit{adapters}, \textit{prompt-based fine-tuning}, and \textit{sparse subnetworks}. \textit{Adapters}~\citep{rebuffi-redidual-adapters-2017, pmlr-v97-houlsby19a} are additional trainable neural modules injected into each layer of the otherwise frozen PTLM, including their variants~\citep{pfeiffer-etal-2021-adapterfusion}, have been adopted in both single-domain~\citep{bapna-firat-2019-simple} and multi-domain~\citep{hung-etal-2022-ds} scenarios. \textit{Sparse subnetworks}~\citep{hu-etal-2022-lora, ansell-etal-2022-composable} reduce the number of training parameters by keeping only the most important ones, resulting in a more compact model that requires fewer parameters for fine-tuning. \textit{Prompt-based fine-tuning}~\citep{li-liang-2021-prefix, lester-etal-2021-power, goswami-etal-2023-switchprompt} reduces the need for extensive fine-tuning with fewer training examples by adding prompts or cues to the input data. These approaches, however, are still parameter- and time-inefficient, as they add additional parameters, require complex training steps, are less intuitive to the expressiveness, or are limited to the multi-domain scenario for domain adaptation. A broader overview and discussion of recent domain adaptation methods in low-resource scenarios is given in the survey of~\citet{hedderich-etal-2021-survey}.

\section{Conclusions}
%\vspace{-0.4em}
In this paper, we introduced TADA -- a novel task-agnostic domain adaptation method which is modular and parameter-efficient for pre-trained transformer-based language models. We demonstrated the efficacy of \TADA in 4 downstream tasks across 14 domains in both single- and multi-domain settings, as well as high- and low-resource scenarios. % without conducting complex training steps or introducing additional parameters during domain-adaptive pre-training. 
An in-depth analysis revealed the advantages of \TADA in few-shot transfer and highlighted how our domain-specialized tokenizers take the domain vocabularies into account. We conducted the first study on meta-tokenizers and showed their potential in combination with meta-embeddings in multi-domain applications. 
Our work points to multiple future directions, including advanced meta-tokenization methods %in multi-domain scenario 
and the applicability of \TADA beyond the studied tasks in this paper.

\section*{Acknowledgements}
We would like to thank the members of the NLP and Neuro-Symbolic AI research group at the Bosch Center for Artificial Intelligence (BCAI) and the anonymous reviewers for their feedback.

\section*{Limitations}
In this work, we have focused on the efficiency concerns of task-agnostic domain adaptation approaches leveraging pre-trained transformer-based language models. The experiments are conducted on four tasks across 14 domains in both high- and low-resource scenarios. We only consider the methods utilizing pre-collected in-domain unlabeled text corpora for domain-adaptive pre-training. It is worth pointing out that the selected domains are strongly correlated to the selected tasks, which does not reflect the wide spectrum of domain interests. Besides, the datasets are covered only in English to magnify the domain adaptation controlling factors and use cases, while multilinguality would be the next step to explore. We experimented on encoder-only PTLM based on the downstream classification tasks, where the encoder-decoder PTLM would be applicable to different tasks (e.g., natural language generation, summarization, etc.) requiring more computational resources.  We hope that future research builds on top of our findings and extends the research toward more domains, more languages, more tasks, and specifically with the meta-tokenizers for efficiency concerns of domain adaptation approaches. 

% ACL 2023 requires all submissions to have a section titled ``Limitations'', for discussing the limitations of the paper as a complement to the discussion of strengths in the main text. This section should occur after the conclusion, but before the references. It will not count towards the page limit.
% The discussion of limitations is mandatory. Papers without a limitation section will be desk-rejected without review.

% While we are open to different types of limitations, just mentioning that a set of results have been shown for English only probably does not reflect what we expect. 
% Mentioning that the method works mostly for languages with limited morphology, like English, is a much better alternative.
% In addition, limitations such as low scalability to long text, the requirement of large GPU resources, or other things that inspire crucial further investigation are welcome.
\section*{Ethics Statement}
%\textcolor{blue}{use the pre-collected background and task, might result in potential risk of ethical statements including biases.}
We utilized the pre-collected in-domain unlabeled text corpora to explore the domain-adaptation pre-training approaches with efficiency concerns in this work. Although we carefully consider the data distribution and the selection procedures, the pre-collected background sets for each domain might introduce the potential risk of sampling biases. Moreover, (pre)training, as well as fine-tuning of large-scale PTLMs, might pose a potential threat to the environment~\citep{strubell-etal-2019-energy}: in light of the context, the task-agnostic domain adaptation approaches we introduced are aimed at mitigating towards the directions of reducing the carbon footprint of pretrained language models.

% Scientific work published at ACL 2023 must comply with the ACL Ethics Policy.\footnote{\url{https://www.aclweb.org/portal/content/acl-code-ethics}} We encourage all authors to include an explicit ethics statement on the broader impact of the work, or other ethical considerations after the conclusion but before the references. The ethics statement will not count toward the page limit (8 pages for long, 4 pages for short papers).

\bibliography{custom}
%\bibliography{anthology.custom}
\bibliographystyle{acl_natbib}

\appendix

\clearpage
\onecolumn
\label{sec:appendix}

\section{Computational Information}
All the experiments are performed on Nvidia Tesla V100 GPUs with 32GB VRAM and run on a carbon-neutral GPU cluster.
The number of parameters and the total computational budget for domain-adaptive pre-training (in GPU hours) are shown in Table \ref{tab:computational}.

\setlength{\tabcolsep}{13.4pt}
\begin{table*}[h]
\centering
\scriptsize{
\begin{tabular}{lll}
\toprule
% \multirow{2}{*}{\textbf{Model}}                     & \multirow{2}{*}{\textbf{\# Parameters}}                                     & \multirow{2}{*}{{\textbf{\begin{tabular}[c]{@{}l@{}}Budget\\ (in GPU hours)\end{tabular}}}}  & \multirow{2}{*}{\textbf{Infrastructure}}        \\ 
\textbf{Model}   & \textbf{\# Trainable Parameters} & \textbf{MLM Budget (in GPU hours)} \\\midrule
BERT$\ddag$ (\mlmFull)  & $\sim$110 M & $\sim$5.5h (NER and NLI), 7.5h (DST and RR) \\
BERT$\ddag$ (\mlmAdapt) & $\sim$0.9 M & $\sim$2.5h (NER and NLI), 3.5h (DST and RR)  \\
BERT$\ddag$ (\mlmEmb)   & $\sim$24 M  & $\sim$3.5h (NER and NLI), 4.5h (DST and RR)  \\
\bottomrule
\end{tabular}
}
\caption{Overview of the computational information for the domain-adaptive pre-training. $\ddag$BERT variants: BERT (NLI, NER) and TOD-BERT (DST, RR).}
\label{tab:computational}
\end{table*}

\section{Hyperparameters}
Detailed explanations of our hyperparameters are provided in the main paper in Section~\autoref{subsec:hyperparameters_and_optimization}. 
In our conducted experiments, we only search for the learning rate in domain-adaptive pre-training. The best learning rate depends on the selected domains and methods for each task.

\section{In-domain Unlabeled Text Corpora}
\label{appendix:domain-corpora}
We provide more detailed information on the background datasets that are used for domain-adaptive pre-training in Table~\ref{tab:background_data}.

\setlength{\tabcolsep}{10.6pt}
\begin{table*}[h]
\centering
\scriptsize{
\begin{tabular}{lllr}
\toprule
\textbf{Task}                     & \textbf{Domain}                                     & \textbf{Background dataset}     & \textbf{\# Sentences}        \\ \midrule
\multirow{5}{*}{DST, RR} & Taxi       &  \multirow{5}{*}{DomainCC corpus from~\citet{hung-etal-2022-ds}.}     & 200 K \\
                         & Restaurant & & 200 K          \\
                         & Hotel      & & 200 K                      \\
                         & Train      & & 200 K                      \\
                         & Attraction & & 200 K                      \\ \midrule
\multirow{5}{*}{NLI}     & Government & \multirow{4}{*}{The respective part of the OANC corpus. }  & 46.0 K                 \\
                         & Travel     & & 47.4 K                               \\
                         & Slate      & & 214.8 K                               \\
                         & Telephone  & & 234.6 K                           \\ \cmidrule{2-4}
                         & Fiction    & The books corpus \citep{zhu2015aligning}, used as the pre-training data of BERT \citep{devlin-etal-2019-bert}.  & 299.5 K                \\\midrule
\multirow{7}{*}{NER}    & \multicolumn{1}{l}{\multirow{2}{*}{News}} & The Reuters news corpus in NLTK (\texttt{nltk.corpus.reuters}). Similar to the  training data  of    & \multicolumn{1}{r}{\multirow{2}{*}{51.0 K}}                                \\
& \multicolumn{1}{c}{}  & CoNLL~\citep{tjong-kim-sang-de-meulder-2003-introduction}. & \\
& Clinical              & Pubmed abstracts from clinical publications filtered following~\citet{lange-etal-2021-clin-x}.  & 299.9 K                \\
                        & Financial         & The financial phrase bank from~\citet{malo2014good}.  & 4.8 K                  \\
                        & Fiction              &  Same as NLI \textsc{Fiction}, described above.   & 299.5 K    \\
                         
              %           & Science              & Randomly sampled SemanticScholar abstracts from Biology (70\%) and Computer Science (30\%). Similar to the pre-training data of SciBERT~\citep{beltagy-etal-2019-scibert}.    & 300.1 K            \\
                          & \multicolumn{1}{l}{\multirow{2}{*}{Science}} & Randomly sampled SemanticScholar abstracts from Biology (70\%) and Computer Science (30\%). & \multicolumn{1}{r}{\multirow{2}{*}{300.1 K}}  \\
 & \multicolumn{1}{c}{} &  Similar to the pre-training data of SciBERT~\citep{beltagy-etal-2019-scibert}. & \\
                         \bottomrule
\end{tabular}%
}
\caption{Overview of the background datasets and their sizes as reported in Table~\ref{tab:data-analysis} in the background column. The background datasets are used to train domain-specific tokenizers and domain-adapted embeddings layer.}
\label{tab:background_data}
\end{table*}

\clearpage

\section{Per-Domain Results for Meta-Tokenizers}
\label{appendix:md-mt}
We provide the results for each domain in our multi-domain experiments with meta-tokenizers and meta-embeddings in Table~\ref{tab:eval_result_multi_metatok} for DST and RR, and in Table~\ref{tab:eval_result_mnli_ner_multi_metatok} for NLI and NER.

\setlength{\tabcolsep}{2.3pt}
\begin{table*}[h]
\centering
\scriptsize{
\begin{tabular}{l ccccc g ccccc g}
\toprule
& \multicolumn{6}{c}{\textbf{DST}}     & \multicolumn{6}{c}{\textbf{RR}}          \\ 
\textbf{Model} & \textbf{Taxi} & \textbf{Restaurant} & \textbf{Hotel} & \textbf{Train} & \textbf{Attraction} & \textbf{Avg.} &  \textbf{Taxi} & \textbf{Restaurant} & \textbf{Hotel} & \textbf{Train} & \textbf{Attraction} & \textbf{Avg.}\\ \cmidrule(lr){2-7}\cmidrule(lr){8-13}
(AVG) TOD-BERT (\mlmEmbs)          & 35.42 & 46.71 & 40.82 & \textbf{52.34}& 47.30 & 44.52 & \textbf{55.20} & \textbf{64.58} & \textbf{60.39} & \textbf{62.84} & \textbf{66.11} & \textbf{61.82} \\
(ATT) TOD-BERT (\mlmEmbs)          & 37.35 &  \textbf{46.98}     & \textbf{41.32}  & 51.92  & \textbf{47.88}  & \textbf{45.09} & 53.73 & 64.00     & 59.89 & 61.54 & 65.05 & 60.84 \\\midrule
(AVG) TOD-BERT (\mlmEmbsTokX) dyn & 32.06 & 44.12 & 40.54 & 49.89 & 44.21 & 42.16 & 52.84 & 62.54 & 58.26  &  61.24 & 64.46 & 59.87 \\
(AVG) TOD-BERT (\mlmEmbsTokX) space & 31.35 & 44.89 & 37.27 & 49.47 & 44.86 & 41.57 & 51.59 & 62.46 & 56.44 & 60.21 & 61.99 & 58.54 \\
(AVG) TOD-BERT (\mlmEmbsTokX) trun & 33.61 & 43.88 & 38.20 & 44.24 & 41.35 & 40.26 & 52.55 & 61.19 & 55.55 & 58.58 & 62.47 & 58.07\\
(ATT) TOD-BERT (\mlmEmbsTokX) dyn & 34.06 & 45.01 & 39.73 & 50.11 & 44.73 & 42.73 & 51.22 & 62.08 & 58.04 & 61.39 & 63.35 & 59.22 \\
(ATT) TOD-BERT (\mlmEmbsTokX) space &30.19 & 42.57 & 40.23 & 49.84 & 44.41 & 41.45 & 51.51 & 61.64 & 57.30 & 60.91 & 63.41 & 58.95\\
%(ATT) TOD-BERT (\mlmEmbsTokX) pad &  &  &  &  &  & &  &  &  &  &  &\\
(ATT) TOD-BERT (\mlmEmbsTokX) trun & 31.45 & 43.44  & 37.08  & 48.13 &  44.02 & 40.82 & 51.59 & 62.63  & 57.97  & 60.66 & 62.62 & 59.09\\
  \bottomrule
\end{tabular}%
}
\caption{Results of meta-tokenizers in multi-domain experiments with meta-embeddings on two downstream tasks: DST and RR, with joint goal accuracy (\%) and $\textsc{R}_{100}@1$ (\%) as evaluation metric, respectively. Three meta-tokenization aggregation methods: dynamic (dyn), space (space), truncation (trun), are combined with two meta-embeddings approaches: average (\textsc{AVG}), attention-based (\textsc{ATT}).}
\label{tab:eval_result_multi_metatok}
%\vspace{-0.5em}
\end{table*}

\setlength{\tabcolsep}{2.6pt}
\begin{table*}[h]
\centering
\scriptsize{
\begin{tabular}{l ccccc g ccccc g}
\toprule
 & \multicolumn{6}{c}{\textbf{NLI}}    & \multicolumn{6}{c}{\textbf{NER}}    \\ 
\textbf{Model} & \textbf{Government} & \textbf{Telephone} & \textbf{Fiction} & \textbf{Slate} & \textbf{Travel} & \textbf{Avg.} & \textbf{Financial} & \textbf{Fiction} & \textbf{News} & \textbf{Clinical} & \textbf{Science} &  \textbf{Avg.}\\ \cmidrule(lr){2-7}\cmidrule(lr){8-13}
(AVG) BERT (\mlmEmbs)         &\textbf{83.80}&	80.87&	81.70&	\textbf{77.60}&	\textbf{81.30}&	\textbf{81.05} & 87.72	&68.78	&90.16&	85.68&	78.22&	82.11 \\
(ATT) BERT (\mlmEmbs)            &83.50&	81.64	&\textbf{81.74}	&76.68	&80.36& 80.78& \textbf{88.89}&	69.05&	\textbf{90.56}&	85.43&	\textbf{80.55}&	\textbf{82.90}\\\midrule
(AVG) BERT (\mlmEmbsTokX) dyn  & 81.08 & 79.81 & 80.44 & 75.35 & 78.80 & 79.10 & 83.26 & 59.70  & 75.93 & 70.42 & 64.33 & 70.73 \\
(AVG) BERT (\mlmEmbsTokX) space & 81.90 & 81.33 & 80.49 & 75.14 & 78.69 & 79.51 & 83.68 & 61.68 & 76.39 & 70.78 & 60.61 & 70.63 \\
(AVG) BERT (\mlmEmbsTokX) trun & 81.44  & 81.38 & 79.17 & 75.86 &  79.50 & 79.47  & 77.99 & 53.53 & 74.37 & 67.08 & 60.33 &  66.66 \\
(ATT) BERT (\mlmEmbsTokX) dyn  & 81.70 & 80.62 & 80.33 & 74.78 & 79.15 & 79.32 & 84.64 & 59.98 & 76.08 & 71.30 & 62.17 & 70.83 \\
(ATT) BERT (\mlmEmbsTokX) space & 83.34 & 81.43 & 80.23 & 74.83 & 79.81 & 79.93 & 83.70 &  62.03 & 76.04 & 71.54 & 60.22 & 70.71\\
(ATT) BERT (\mlmEmbsTokX) trun  & 82.37 & 81.64 & 78.81 & 75.65 & 79.90 & 79.67 & 80.33& 58.80 & 74.49 & 66.92 & 61.51 & 68.41  \\
  \bottomrule
\end{tabular}%
}
\caption{Results of meta-tokenizers in multi-domain experiments with meta-embeddings on two downstream tasks: NLI and NER, with accuracy (\%) and $F_1$ (\%) as the evaluation metric, respectively. Three meta-tokenization aggregation methods: dynamic (dyn), space (space), truncation (trun), are combined with two meta-embeddings approaches: average (\textsc{AVG}), attention-based (\textsc{ATT}).}
\label{tab:eval_result_mnli_ner_multi_metatok}
%\vspace{-0.5em}
\end{table*}

\end{document}